\documentclass[10pt,conference]{IEEEtran}
\IEEEoverridecommandlockouts
\usepackage{cite}
\usepackage{tabularx,booktabs}
\usepackage{amsmath,amssymb,amsfonts}
\usepackage{algorithmic}
\usepackage{graphicx}
\usepackage{textcomp}
\usepackage{xcolor,colortbl}
\def\BibTeX{{\rm B\kern-.05em{\sc i\kern-.025em b}\kern-.08em
    T\kern-.1667em\lower.7ex\hbox{E}\kern-.125emX}} 

\definecolor{Gray}{gray}{0.85}
\newcolumntype{a}{>{\columncolor{Gray}}c}

\DeclareUnicodeCharacter{2212}{-}

\author{\IEEEauthorblockN{Huy T. Nguyen\IEEEauthorrefmark{1}\IEEEauthorrefmark{3}, Thinh B. Lam\IEEEauthorrefmark{2}, Quan T. D. Tran\IEEEauthorrefmark{3}, Minh T. Nguyen\IEEEauthorrefmark{4}, Dat T. Chung\IEEEauthorrefmark{2} and Vinh Q. Dinh\IEEEauthorrefmark{6}} 
}

\title{In-context Cross-Density Adaptation on Noisy Mammogram Abnormalities Detection

\thanks{
\IEEEauthorrefmark{7}This paper is partially supported by AI VIETNAM.\\
\IEEEauthorrefmark{1}VinBrain JSC, Vietnam \\
\IEEEauthorrefmark{2}VinBigData, Vietnam \\
\IEEEauthorrefmark{3}University of Education, Vietnam \\
\IEEEauthorrefmark{4}University of Medicine and Pharmacy, Vietnam \\
\IEEEauthorrefmark{6}Vietnamese-German University, Vietnam}
}

\begin{document}
\maketitle
\begin{abstract}
This paper investigates the impact of breast density distribution on the generalization performance of deep-learning models on mammography images using the VinDr-Mammo dataset. We explore the use of domain adaptation techniques, specifically Domain Adaptive Object Detection (DAOD) with the Noise Latent Transferability Exploration (NLTE) framework, to improve model performance across breast densities under noisy labeling circumstances. We propose a robust augmentation framework to bridge the domain gap between the source and target inside a dataset. Our results show that DAOD-based methods, along with the proposed augmentation framework, can improve the generalization performance of deep-learning models (+5\ \% overall mAP improvement approximately in our experimental results compared to commonly used detection models). This paper highlights the importance of domain adaptation techniques in medical imaging, particularly in the context of breast density distribution, which is critical in mammography screening.

\end{abstract}


\section{Introduction}


Regular breast cancer screening is essential for detecting tumors early. Computer-aided diagnosis (CADx) systems incorporate mammography, an X-ray imaging technique, to help radiologists improve their efficiency in detecting breast cancer. Research studies such as \cite{risk1} and \cite{risk2} have established a strong link between high breast density and an increased risk of developing breast cancer. It shows that women have up to four to six times higher risk compared to those with low breast density. As a result, mammograms with high breast density are more common than those with low breast density, the latter of which is even less common in cases of cancer.  


Many deep convolutional networks (CNN)-based models \cite{sam1},\cite{sam2} achieved excellent performance on some well-known mammogram datasets such as VinDr-Mammo\cite{vindr}, INbreast\cite{inbreast} or CBIS-DDSM\cite{ddsm} \. Having said that, they often struggle to adapt to the highly diverse medical environments of mammograms, such as varying backgrounds, patients' specialized features, and different devices used for screening. It may cause a significant domain shift between the training and test data. 


In mammograms, many works tackle breast cancer screening using this domain adaptation. Wang et.al.\cite{wang} proposed a domain adaptation method based on deep adversarial learning to conduct knowledge transfer from the public dataset to their target dataset. In the object detection task, Domain Adaptive Faster Region-based Convolutional Neural Networks (DA-FRCNN) was the first Domain Adaptation Object Detection (DAOD) proposed by Chen et.al.\cite{dafrcnn} to address domain shift problems. Several latest studies developed more superior frameworks using the domain adaptation mechanism\cite{strongweak}, \cite{gpa}, outperforming results in adaptation from many popular datasets such as Sim10k to Cityscape or PASCAL to Clipart dataset.

\begin{figure*}[t]
	\centering
	\includegraphics[width=0.8\linewidth]{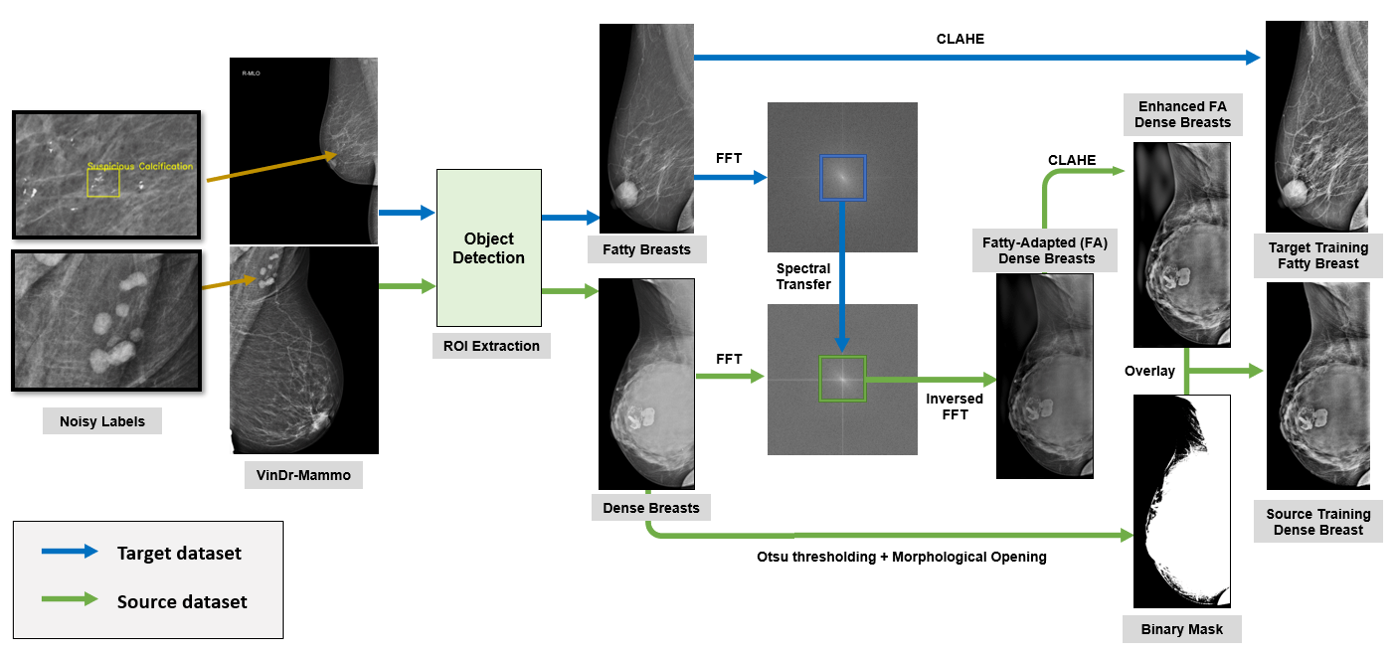}
	\caption{FALCE: Fourier-adapted Locality Contrast Enhancement framework for preprocessing and creating the fatty-adapted dense images before applying CLAHE under DIP-based created a binary mask for enhancing locality semantic information.}
	\label{fig: baseline compare}
\end{figure*}

Current methods for medical image analysis are typically designed for clean source domains, which is not always practical in real-world applications. In the medical domain, noisy class annotations are common due to the high level of expertise required of annotators. This can lead to issues such as high annotating costs and time requirements, making noisy labeling a significant challenge for medical image analysis. Liu et al.\cite{NLTE} proposed a novel Noise Latent Transferability Exploration (NLTE) framework to address the negative effects of noisy labeling caused by incorrectly annotated and class-corrupted samples. 

Several data augmentation strategies have been proposed to boost training performance such as Generative Adversarial Network-based medical image synthesis in breast screening \cite{gan}. Fourier Domain Adaptation (FDA)\cite{FDA} has outlined a straightforward technique for unsupervised domain adaptation that reduces the disparity between the source and target distributions by switching the low-frequency spectra of the two. 


In this paper, we first design a new robust augmentation framework for improving cross-domain detection performance in the designed context. We introduce a local transformation-based framework that focuses on the informative areas of breast screening, then applied contrast-enhanced domain shift information from the target domain to the source dataset.

Secondly, we tackle one of the latest and biggest breast screening datasets - VinDr-Mammo \cite{vindr} on the abnormalities detection task. We split VinDr-Mammo into the Fatty breast sub-dataset and the Dense breast sub-dataset. The redesigned NLTE framework with regularization-added DAOD is brought for further assessment of evaluating the noisy effect of breast screening annotation. Our proposed method performed significantly better than commonly used object detection methods.

\section{Methodology}

\subsection{Mammogram Region of Interest Extraction}

In practice, radiologists perform X-ray examinations of the patient's breasts containing a substantial black area with no useful information. It causes a high training computational cost and affects the performance of our proposed augmentation method through global mammogram distribution. Therefore, a light one-stage object detection model is brought to capture the relevant region of the breast accurately from the original scans. The Region Of Interest (ROI) extraction model is trained by using active learning, we manually labeled 1000 images in total. Labeled scans containing four views, then 800 images were fed for the YOLOv7 \cite{yolov7} model, and validated on 200 images. The model trained on the above initial set of labeled instances is used to make predictions on the other 1000 scans before validating and adjusting labels for the next phase of training. While ensuring that the most informative occurrences are labeled, active learning strategy help to reduce the time and effort needed for manual labeling. Finally, the trained model was able to accurately estimate the bounding boxes of the breast in observed mammograms and obtained mean Average Precision (mAP) of 0.9985 on the validation set.

\subsection{Fourier-adapted locality contrast enhancement (FALCE)}

\begin{figure*}[t]
	\centering
	\includegraphics[width=0.85\linewidth]{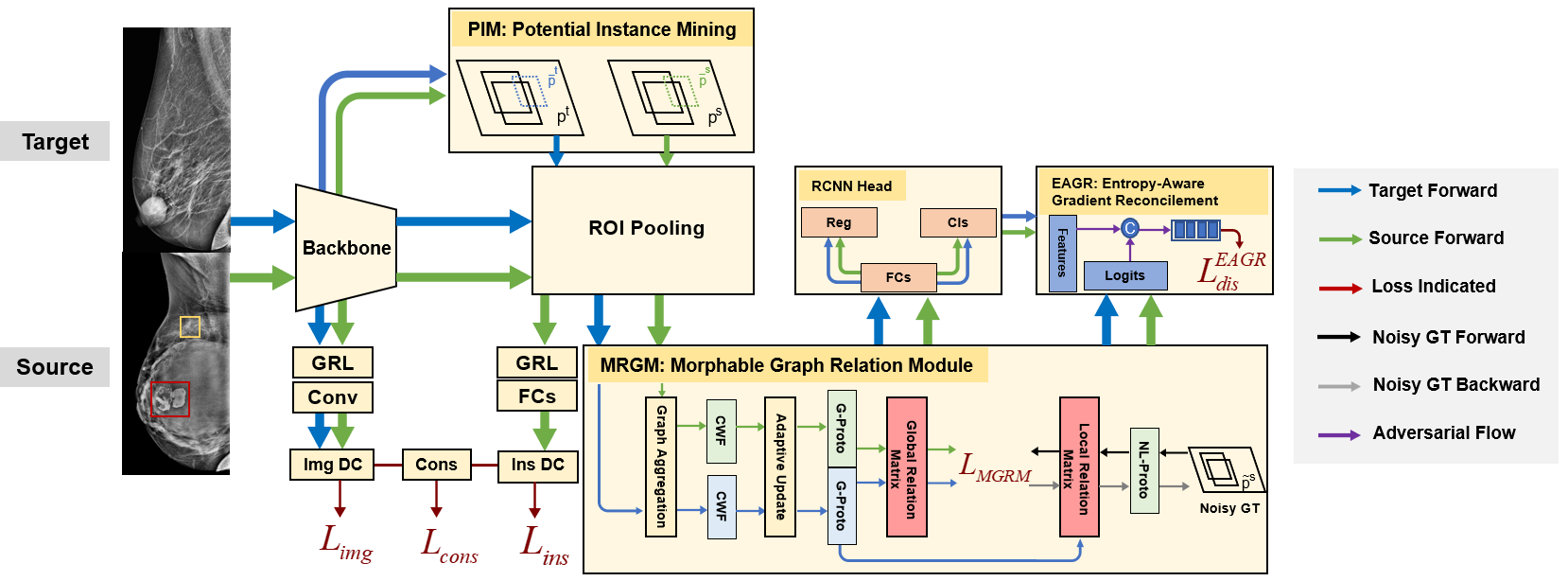}
	\caption{Overview of redesigned loss Noise Latent Transferability Exploration for adapting and noisy handling source dense breasts and target fatty breasts.}
	\label{fig: baseline compare}
\end{figure*}

Domain shift problems between two breast dataset distributions can be handled in preprocessing stage or training stage. In this study, a novel framework called Fourier-adapted locality contrast enhancement is proposed for applying the local transformation on mammograms for mapping two distribution spaces closer. Inspired by FDA\cite{FDA} and Contrast Limited Adaptive Histogram Equalization (CLAHE) \cite{CLAHE}, the framework aims to conduct spectral transfer, mapping a source image to a target without changing semantic content. Mathematically, given that ${F^A},\,{F^P}:{R^{H \times W \times 1}} \to {R^{H \times W \times 1}}$ are the amplitude and phase components of the Fourier transform $F$ of a mammogram, we have
\begin{equation}
  \resizebox{0.8\hsize}{!}{$F(x)(m,n) = \sum\limits_{h,w} {x(h,w){e^{ - k2\pi \left( {\frac{h}{H}m + \frac{w}{W}n} \right)}}}$},
\end{equation}
where $k^2 = -1$. $F^{-1}$ is the inverse Fourier transform mapping spectral information back to 2D-image space. With mask $M_{\beta}$ contains zero value except center region with $\beta \in (0,1)$ as
\begin{equation}
  {M_\beta }(h,w) = {\Im _{(h,w) \in \left[ { - \beta H:\beta H, - \beta W:\beta W} \right]}},
\end{equation}
Sampled images $x^s \sim D^s$,$x^t \sim D^t$ are taken randomly, FDA is shown as
\begin{equation}
 {x^{s \to t}} = {F^{ - 1}}\left( {{M_\beta } \circ {F^A}({x^t}) + \left( {1 - {M_\beta }} \right) \circ {F^A}({x^s}),{F^P}({x^s})} \right),
\end{equation}

After that, the fatty-adapted dense breasts are passing through CLAHE for enhancing the global information for the detection model. However, the enhanced FA Dense Breasts with noise in the background might lower the overall performance. The original dense breasts are brought to create a binary mask by using Otsu thresholding combined with the opening operator. Given that the morphological opening of an image $f$  by a structuring element s (denoted by $f \circ s$) is an erosion followed by a dilation: $f \circ s = \left( {f\ominus B} \right) \oplus B$, where $\oplus,\ominus$ are dilation and erosion operation, respectively. Opening tends to smooth contours in the mammograms, remove isolated bright points, and break tenuous connections between regions in the image~\cite{Opening}. The output of the framework is an overlayed image of a created binary mask and its corresponding enhanced FA Dense Breasts. The fatty breast images also are applied CLAHE before we use them for the training phase.

\subsection{Domain Adaptive Object Detection with NLTE framework}

The domain gaps between labeled source data and unlabeled target data have been narrowed using unsupervised domain adaptive object detection. In this study, the detection model is built based on DA-FRCNN and consisted of 2 parts: Faster-RCNN and Domain Adaptive Components. Faster R-CNN is a two-stage detector that primarily consists of three key elements: a region proposal network (RPN), a region-of-interest (ROI) based classifier, and shared bottom convolutional layers. In this NLTE framework, the ROI part is redesigned as Potential Instance Mining (PIM) for recapturing potential  foreground instances from the background as follows:
\begin{equation}
  \overline{{{\bf P}}}^{s}=\{\overline{{{p}}}_{i}\mid o(\overline{{{p}}}_{i}) >\tau,\overline{{{p}}}_{i}\not\in{\bf P}^{s},\forall {p_j}\in {\bf P}^{s}:IoU(\overline{{{p}}}_{i},p_{j})=0 \},
\end{equation}
where $\overline{{{\bf P}}}^{s}$ and ${\bf P}^{s}$ are features from PIM and original proposals generated by RPN,  respectively. $o(\overline{{{p}}}_{i})$ is objectness score as the output of RPN, $\tau$ is the threshold, and $IoU$ is Intersection over Union. PIM is used for both source and target domains. Only highly confident proposals are kept through the PIM method, ensuring that missing items are recaptured.  This improves the number of correctly labeled examples for improving discriminating ability and enriches the diversity of source semantic properties. 


The component loss of the base model is the sum of PIM and ROI classifier loss is $\mathcal{L}_{det} = \mathcal{L}_{pim} + \mathcal{L}_{roi}.$ The RPN and ROI classifiers both include two loss terms for their training data: one for classification, which measures how accurately the predicted probability is predicted, and the other, which is a regression loss on the box coordinates for improved localization.

To synchronize the feature representation distributions on those two differences of breast domains, domain adaptation components for the images and instance levels are proposed. Denoting that $D_i$ is the domain label of $i^{th}$ training image, where $D_i = 0$ and $D_i = 1$ stand for source and target domain, respectively. Cross entropy loss is used for the image-level loss with the output of the domain classifier $p_i$ defined as
\begin{equation}
  \mathcal{L}_{img} = - \mathop \sum \limits_i \left[ {{D_i} \cdot \log ({p_i}) + (1 - {D_i}) \cdot \log (1 - {p_i})} \right].
\end{equation}
To align the instance-level distribution, we train a domain classifier for the feature vectors in a manner similar to the image-level adaption. The result of the instance-level domain classifier for the $i^{th}$ image's $j^{th}$ area proposition will be denoted as $p_{i,j}$. Now, it is possible to express the instance-level adaption loss as
\begin{equation}
  \mathcal{L}_{ins} = - \mathop \sum \limits_{i,j} \left[ {{D_i} \cdot \log ({p_{i,j}}) + (1 - {D_i}) \cdot \log (1 - {p_{i,j}})} \right].
\end{equation}
For balancing and strengthening two branches of image-level and instance-level adaptation, we added consistency regularizer term  to be enforced with loss written as
\begin{equation}
  \mathcal{L}_{cons} = {\sum\limits_{i,j} {\left\| {\frac{1}{{\left| I \right|}}\sum {{p_i}}  - {p_{i,j}}} \right\|} _2},
\end{equation}
where $I$ stands for the number of activations in a feature map, and $||.||_2$ is Euclidean distance.

At the bottom of Fig 2., the Morphable Graph Relation Module is conducted for simulating the adaptability and transition probability of class-corrupted samples. It examines the underlying domain knowledge and semantic information inside these samples. Using morphable graphs, the category-wise relationships are regularized between noisy local prototypes and global prototypes. After that, Global Relation Matrix (GRM) is synthesized which is updated by the aggregated node features to model the class-wise transition probability in the context of domain crossing. Local relation matrices (LRM) are constructed to investigate the alignment viability of noisy data, and the GRM regularizes the transition probabilities of noisy samples. To regularize the transition probabilities between the local relation matrix and the global relation matrix, we employ the $l_1$ loss defined as 
\begin{equation}
 {L_{mgrm}} = \frac{1}{r}\sum\limits_{z \in \Im (Z)} {\left| {{z_r} - {\pi _r}} \right|,} 
\end{equation}
where $z_r$, $\pi _r$, respectively, are LRM and GRM. $\Im (Z)$ is the non-zero columns within $Z$, showed that the $r^{th}$ category exists within the batch. This framework applied MGRM to achieve effective semantic alignment between source and target knowledge.

The domain adaptive detector plays a min-max game to produce a saddle-point solution, thereby reducing the target domain's performance degradation $\left( {{{\hat \phi }_f},{{\hat \phi }_{\det }},{{\hat \phi }_{dis}}} \right)$ as

\begin{equation}
\,\,\,\, \left( {{{\hat \phi }_f},{{\hat \phi }_{\det }}} \right) = \arg \left( {\mathop {\min }\limits_{{\phi _f},{\phi _{\det }}} ({L_{\det }} - {L_{dis}})} \right),
\end{equation}

and

\begin{equation}
\left( {{{\hat \phi }_{dis}}} \right) = \arg \left( {\mathop {\min }\limits_{{\phi _{dis}}} {L_{dis}}} \right), \,\,\,\,\,\,\,\,\,\,\,\,
\end{equation}
where ${{\hat \phi }_f}$, ${{\hat \phi }_{\det }}$, and ${{\hat \phi }_{dis}}$ stand for optimal parameters of the feature extractor, the detector, and the discriminator, respectively. The whole DAOD framework ${\hat \phi }$ is the combination of those optimal parameters. Similar to the domain classifier branch in the traditional domain adaptation technique, the loss function of the discriminator is written as:

\begin{equation}
L_{dis}^{EAGR} =  - \sum\limits_{i,j} {z\log \left( {\phi _{dis}^{EAGR}\left( {p_i^s \copyright \eta _i^s} \right)} \right)} \,\,\,\,\,\,\,\,\,\,\,\,\\
\end{equation}
 \[\,\,\,\,\,\,\,\,\,\,\,\,\,\, + (1 - z)\log \left( {\phi _{dis}^{EAGR}\left( {p_j^t \copyright \eta _j^t} \right)} \right),\]
where $p_i^s \in P^s$, $p_i^t \in P^t$, $\eta_j^s \in N^s$, $\eta_j^t \in N^t$ are, respectively, source feature, target feature, source corresponding logits, target corresponding logits. $z$ is known as the domain label, with a class source value of 1 and a target value of 0, $\copyright$ is a concatenation operation.

Finally, the following objective function is defined as
\begin{equation}
L = L{ _{\det }} + {\lambda _1}\left( {\sum\limits_{i \in \zeta } {L_{dis}^i\,} } \right) + {\lambda _2}{L_{mgrm}} + L_{dis}^{EAGR},
\end{equation}
where $L{ _{\det }}$ refers to detection loss of base Faster-RCNN model. $\zeta$ is Domain adaptive components space, whereas $\lambda_1$ and $\lambda_2$ are trade-off parameters to balance the effect of the domain adaptation branches. EAGR proposed in NLTE framework to perform meta update to achieve Gradient Reconcile. The adaption process of noisy and clean samples is balanced by this tactic. It then drives the gradients of clean and noisy samples to be consistent toward a domain-invariant direction and associates class confidence with the discriminator.

\section{Experimental results}

\subsection{Dataset preparation}\label{AA}

In this study, the VinDr-Mammo\cite{vindr} dataset is used to conduct the experiments. The dataset contains 20000 images in total, including 1768 and 18232 images recorded with finding and no finding annotations respectively. In the finding images, we divided them into two sets: Dense Density Breasts (DenB) are densities C and D, and Fatty Density Breasts (FatB) are densities A and B. Our experienced doctor verified and preprocessed the dataset, the cleaned DenB and FatB contain 1499 images and 181 images. DenB set is used for training with annotation as the source dataset. We stratified k-fold split  FatB with a ratio of 6:4 into the target training dataset and target test dataset. Due to a lack of samples, there are 4 classes of annotations used for training including Mass (MS), Suspicious Calcification (SC), Asymmetry (AS), and Suspicious Lymph Nodes (LN). AS is a major class of global asymmetry, focal asymmetry, and asymmetry sub-classes released in original data. In conclusion, after our clinician has cleaned the noisy target test dataset, the models are trained on the source dataset (annotated DenB) and target training dataset (FatB without annotations for domain adaptation).
\subsection{Detailed Training \& Evaluation Metrics}

For all of our experiments, we used the same architecture (ResNet-50 \cite{resnet}) as the backbone for the Feature Extractor part of the Detector. The model was trained for 10 epochs using SGD optimizer \cite{sgd} with an initial learning rate $1 \times 10^{−3}$ and decays by $0.1$ after $5$ and $6$ epochs. We resized the shorter side of the image to 640 in both the training and testing process. Our study was built on Pytorch version 1.9.1 and conducted on a machine with NVIDIA RTX 3090Ti GPU. For evaluation, we used the mean Average Precision (mAP) metric on 4-class findings.

\subsection{Experimental Results}

\begin{table}[t]
\caption{Quantitative results (\%) using mAP on the private test dataset}
\label{table: result}
\resizebox{\columnwidth}{!}{%
\begin{tabular}{c|c|c|c|c|a}
    \toprule
    Methods              & MS       & SC    & AS    & LN    & mAP    \\ 
    \midrule 
    Yolov7x              & 72.2     & 17.4  & 28.6  & 63.8  & 45.5   \\ 
    Faster-RCNN          & 71.60	& 17.00	& 36.30	& 25.70	& 37.70  \\ 
    Libra-RCNN           & 66.90	& 11.60 & 48.30	& 29.40	& 39.00  \\ 
    Sparse-RCNN          & 69.80	& 21.80	& 46.20	& 59.70	& 49.40  \\ 
    NLTE                 & 66.88	& 30.26	& 44.49	& 76.36	& 54.50  \\ 
    CLAHE                & 71.19	& 34.19	& 33.53	& 81.82	& 55.18  \\ 
    FDA                  & 70.40	& 36.16	& 33.10	& 80.68	& 55.09  \\ 
    FDA+CLAHE            & 70.53	& 30.24	& 32.71	& 88.92	& 56.83  \\ 
    FALCE (Ours)         & 71.42	& 38.35	& 33.83	& 89.97	& \textbf{57.96}  \\ 
    \bottomrule
\end{tabular}
}
\end{table}

Table \ref{table: result} illustrates the experimental results of the proposed method and the traditional object detectors. It is shown that the object detector with domain adaptation performed better in terms of mAP than traditional object detectors. Specifically, the NLTE object detector improved mAP by $5.1\%$ compared to the best traditional object detector, Sparse-RCNN \cite{sparse}. In addition, NLTE significantly improved the performance of 2 classes (SC and LN with 30.26\% and 76.36\% respectively). 

Based on the superior results of the NLTE method compared to traditional methods, we applied additional augmentation methods and combined them with the NLTE method to boost performance. According to the lower half of Table \ref{table: result}, it shows that the application of augmentation methods improved performance, with our proposed augmentation (FALCE) method achieving the highest possible performance (mAP \textbf{57.96\%}) and improved the performance on 3 classes (MS, SC, and LN with 71.42\%, 38.35\%, and 89.97\%), indicating that our proposed augmentation is robust for the NLTE method. However, the NLTE framework is based on Faster-RCNN, which has some limitations when capturing hard-case patterns like AS, achieving slightly lower Sparse R-CNN and Libra-RCNN.

\section{Conclusion}

In this paper, we proposed a novel unsupervised domain adaptation for Mammogram Abnormalities Detection. The experimental results show that the proposed method significantly outperformed popular object detection methods. As the dataset is small, we plan to investigate and exploit pre-trained models from large datasets, such as ImageNet in our future work.

\vspace{12pt}
\color{red}

\bibliography{IEEEexample}

\begin{thebibliography}{00}

\bibitem{risk1} Boyd, Norman F., et al., "Mammographic density and the risk and detection of breast cancer." New England journal of medicine 356.3 (2007): 227-236.
\bibitem{risk2} Yaghjyan, Lusine, et al., "Mammographic breast density and breast cancer risk: interactions of percent density, absolute dense, and non-dense areas with breast cancer risk factors." Breast cancer research and treatment 150 (2015): 181-189.
\bibitem{sam1} Nguyen, Huyen TX, et al., "A novel multi-view deep learning approach for BI-RADS and density assessment of mammograms." 2022 44th Annual International Conference of the IEEE Engineering in Medicine \& Biology Society (EMBC), IEEE, 2022.
\bibitem{sam2} Tran, Sam B., et al., "Transparency strategy-based data augmentation for BI-RADS classification of mammograms." arXiv preprint arXiv:2203.10609, 2022.
\bibitem{vindr} Nguyen, Hieu Trung, et al., "VinDr-Mammo: A large-scale benchmark dataset for computer-aided diagnosis in full-field digital mammography." MedRxiv (2022): 2022-03.
\bibitem{inbreast}Moreira, Inês C., et al. "Inbreast: toward a full-field digital mammographic database." Academic radiology 19.2 (2012): 236-248.
\bibitem{ddsm} Lee, Rebecca Sawyer, et al. "A curated mammography data set for use in computer-aided detection and diagnosis research." Scientific data 4.1 (2017): 1-9.
\bibitem{wang} Wang, Yan, et al., "Deep adversarial domain adaptation for breast cancer screening from mammograms." Medical image analysis 73 (2021): 102147.
\bibitem{dafrcnn} Chen, Yuhua, et al., "Domain adaptive faster r-cnn for object detection in the wild." Proceedings of the IEEE conference on computer vision and pattern recognition. 2018.
\newpage
\bibitem{strongweak} Saito, Kuniaki, et al., "Strong-weak distribution alignment for adaptive object detection." Proceedings of the IEEE/CVF Conference on Computer Vision and Pattern Recognition, 2019.
\bibitem{gpa} Xu, Minghao, et al., "Cross-domain detection via graph-induced prototype alignment." Proceedings of the IEEE/CVF Conference on Computer Vision and Pattern Recognition, 2020.
\bibitem{NLTE} Liu, Xinyu, et al., "Towards robust adaptive object detection under noisy annotations." Proceedings of the IEEE/CVF Conference on Computer Vision and Pattern Recognition, 2022.
\bibitem{gan}Guan, Shuyue, and Murray Loew, "Breast cancer detection using synthetic mammograms from generative adversarial networks in convolutional neural networks." Journal of Medical Imaging 6.3 (2019): 031411-031411.
\bibitem{FDA} Yang, Yanchao, and Stefano Soatto, "Fda: Fourier domain adaptation for semantic segmentation." Proceedings of the IEEE/CVF Conference on Computer Vision and Pattern Recognition. 2020.
\bibitem{yolov7} Wang, Chien-Yao, Alexey Bochkovskiy, and Hong-Yuan Mark Liao, "YOLOv7: Trainable bag-of-freebies sets new state-of-the-art for real-time object detectors." arXiv preprint arXiv:2207.02696, 2022.
\bibitem{CLAHE} K. Zuiderveld, Contrast Limited Adaptive Histogram Equalization, Graphics gems IV. Academic Press Professional, Inc., USA, 474–485. 1994
\bibitem{Opening} Rafael C. Gonzalez, Richard E. Woods, “Digital Image Processing”, 2nd ed., Beijing: Publishing House of Electronics Industry, 2007.
\bibitem{resnet} He, Kaiming, Zhang, Xiangyu, Ren, Shaoqing and Sun, Jian., "Deep Residual Learning for Image Recognition," IEEE Conference on Computer Vision and Pattern Recognition, 2016.

\bibitem{sgd} Ruder S, "An overview of gradient descent optimization algorithms." arXiv preprint arXiv:160904747, 2016.

\bibitem{sparse} Sun, Peize, et al., "Sparse R-CNN: End-to-End Object Detection with Learnable Proposals." arXiv arXiv:201112450, 2020.

\end{thebibliography}
\end{document}